\documentclass[runningheads]{llncs}
\usepackage{graphicx}
\usepackage{comment}
\usepackage{booktabs}
\usepackage{amsmath,amssymb} % define this before the line numbering.
\usepackage{color}
\usepackage{xcolor}
\usepackage{cite}
\usepackage{subcaption}
\usepackage{gensymb}
\usepackage{microtype}

\begin{document}
\pagestyle{headings}
\mainmatter

\def\ACCV20SubNumber{618}  % Insert your submission number here

%===========================================================
\title{Project to Adapt: Domain Adaptation for Depth Completion from Noisy and Sparse Sensor Data} % Replace with your title
\authorrunning{Lopez-Rodriguez et al.}
\titlerunning{Project to Adapt}

\author{Adrian Lopez-Rodriguez\inst{1}\and 
Benjamin Busam\inst{2, 3}\and 
Krystian Mikolajczyk\inst{1}}

\institute{ 
Imperial College London\\
\email{\{al4415, k.mikolajczyk\}@imperial.ac.uk}\\\and
Huawei Noah's Ark Lab\\\and
Technical University of Munich\\
\email{b.busam@tum.de}}

\maketitle

%===========================================================
%%%%%%%%% BODY
%%%%%%%%% ABSTRACT
\begin{abstract}
%Fusing sparse LiDAR measurements and RGB images is a promising direction to densely estimate depth.
Depth completion aims to predict a dense depth map from a sparse depth input. The acquisition of dense ground truth annotations for depth completion settings can be difficult and, at the same time, a significant domain gap between real LiDAR measurements and synthetic data has prevented from successful training of models in virtual settings. We propose a domain adaptation approach for sparse-to-dense depth completion that is trained from synthetic data, without annotations in the real domain or additional sensors. Our approach simulates the real sensor noise in an RGB~+~LiDAR set-up, and consists of three modules: simulating the real LiDAR input in the synthetic domain via projections, filtering the real noisy LiDAR for supervision and adapting the synthetic RGB image using a CycleGAN~\cite{zhu2017unpaired} approach. We extensively evaluate these modules against the state-of-the-art in the KITTI depth completion benchmark, showing significant improvements.% by 6.4\% in RMSE and 6.3\% in MAE.
\end{abstract}

%%%%%%%%% BODY TEXT
%% Change-log: 1) Bigger fonts Fig. 1/2 (Done, also done for Fig. 4 and 6). 2) Add \cite{wong2020unsupervised}, which is SOTA after ICRA20 (first draft done, need to change % gains claims in the paper) 3) Justification of dataset used, why no interior data (draft in the beginning of Experiments section). 4) PandaSet (https://scale.com/open-datasets/pandaset) is an autonomous driving dataset with RGB + LiDAR setup (no GT). Will add a section testing qualitatively our model only trained in syntethic data directly on that dataset, highlighting both robustness to different camera setups (as we train using a KITTI setup and test in another one) and zero-shot transfer capabilities (we do not need no target data images contrary to self-sup). This was highlighted by the AC. (New figure and paragraph added at the end of the experiments section). 5) Maybe change word "sequences" for the multicamera images as that seemed to confuse Reviewer 3 (he stated VKITTI has 5 sequences, whereas it rather has 5 episodes or scenes, so he thinks our sequences mean scenes/episodes probably). 6) Add number of parameters of networks in Table 2? (ECCV R3, already done) 7) Integrate Table 4 in Table 3 (mentioned by a reviewer, done). 8) Added "+ Proj. + No RGB" in Table 1, need to explain and link to results in Table 2. 9) Probably need to reformulate contributions slightly 9) Add line explaining that CycleGAN changes are mainly contrast/colors. 10) Change title, less obvious re-submission.
\section{Introduction}
\begin{figure*}[t!]
    \centering
    \begin{subfigure}[t]{0.5\textwidth}
        \centering
        \includegraphics[width=0.96\linewidth]{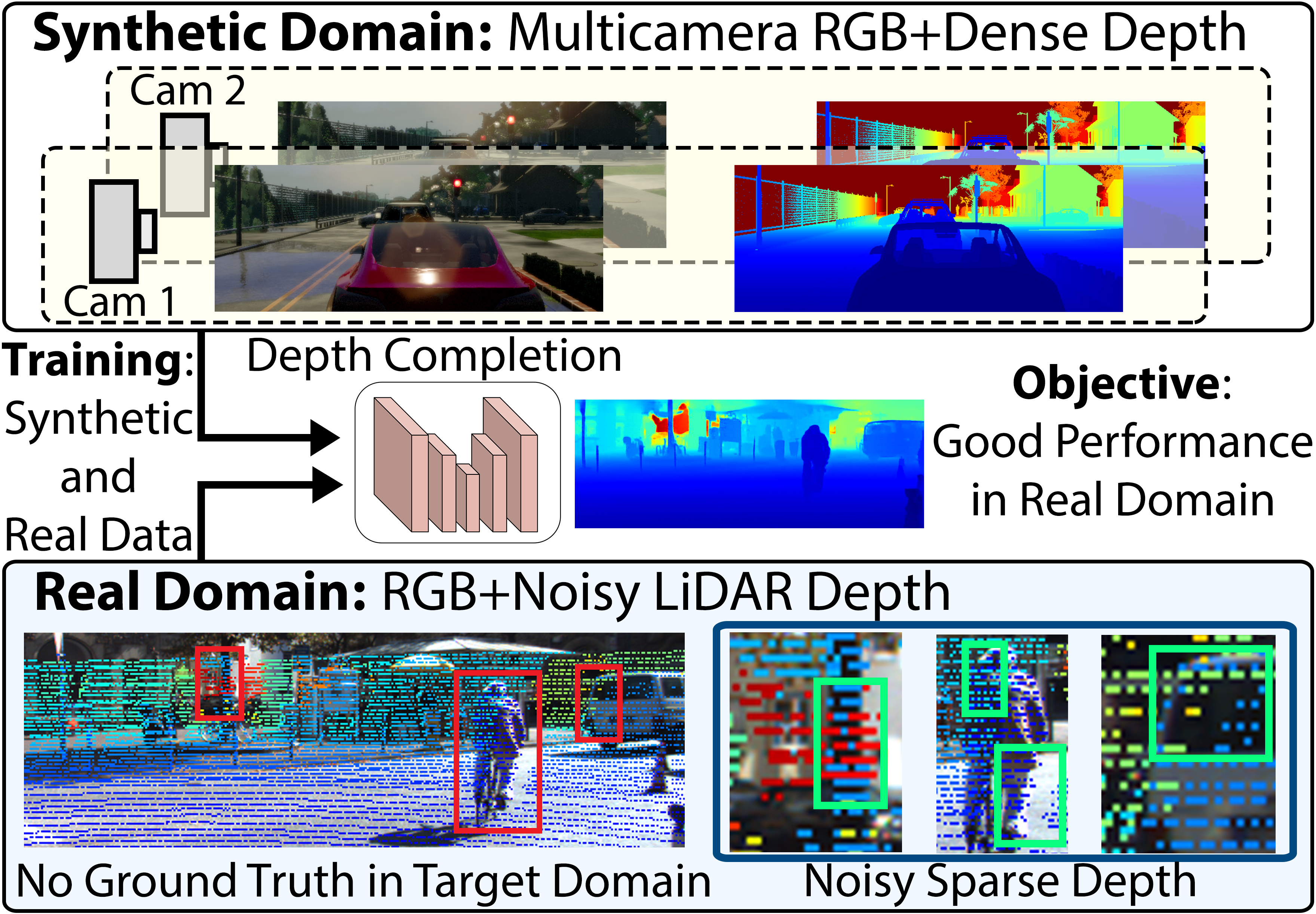}
        \caption{Overview of the approach}
    \end{subfigure}%
    ~ 
    \begin{subfigure}[t]{0.5\textwidth}
        \centering
        \centerline{
        \includegraphics[width=\linewidth,clip]{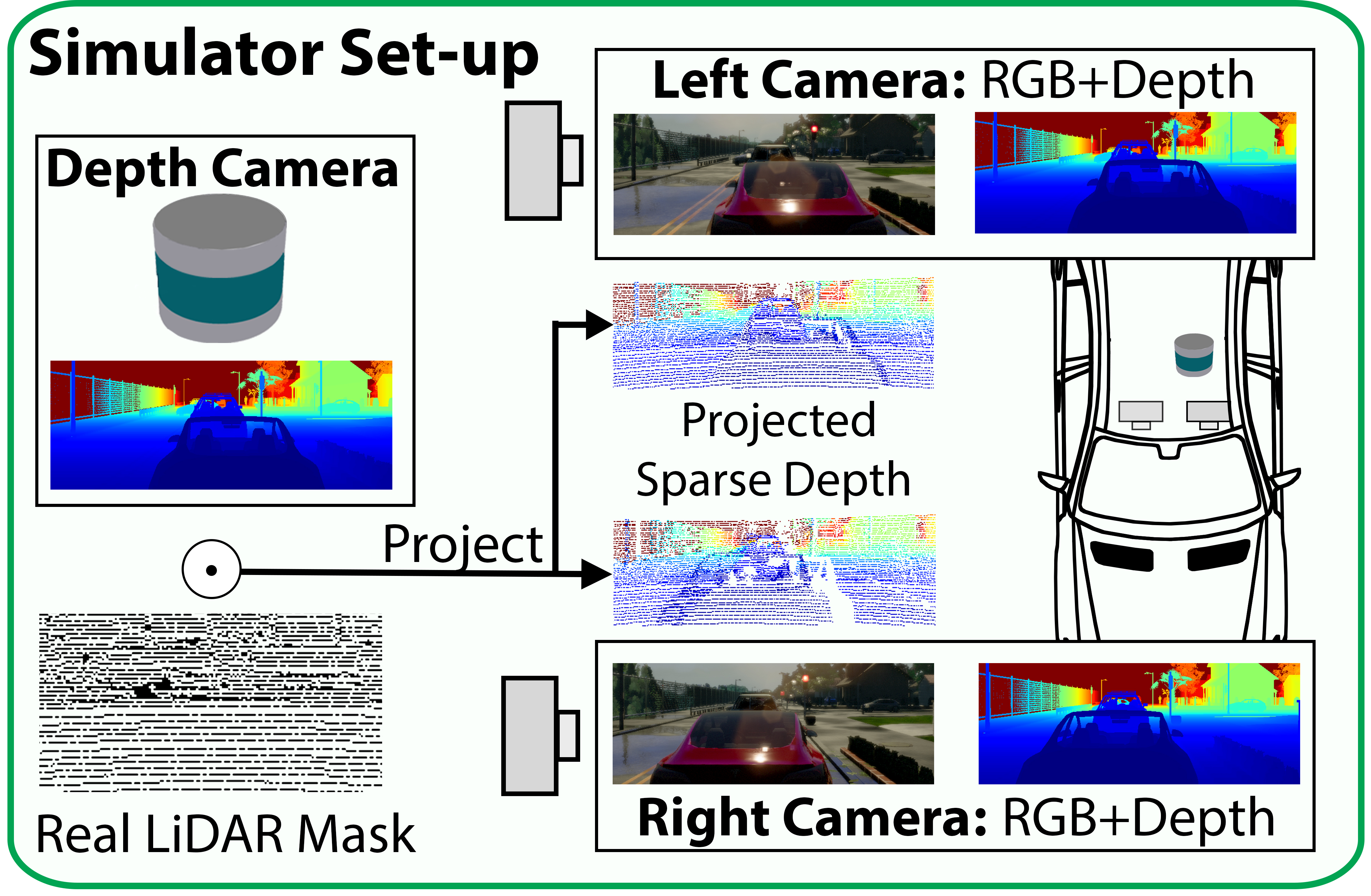}}
        \caption{Multicamera set-up in CARLA}
    \end{subfigure}
    \caption{We investigate the depth completion problem without ground truth annotations in the real domain, which contains paired noisy and sparse depth measurements and RGB images. We highlight some noise present in the real data: see-through on the tree trunk and bicycle, self-occlusion on the bicycle and missing points on the van. In our approach, we leverage synthetic data with multicamera dense depth and RGB images. An overview of the multicamera set-up in CARLA used to simulate the real projection LiDAR artifacts is included. The \textit{Depth Camera} acts as a virtual LiDAR and collects a dense depth, which is sparsified using real LiDAR binary masks and projected to either the \textit{Left Camera} or \textit{Right Camera} reference frame. Both the \textit{Left Camera} and the \textit{Right Camera} collect RGB information, used as part of the input data, and a dense depth map, used for supervision}
    \label{fig:initial}
\end{figure*}

\noindent\textbf{Motivation.} Active sensors such as LiDAR determine the distance of objects within a specified range via a sparse sampling of the environment whose density decreases quadratically with the distance. RGB~cameras densely capture their field of view, however, monocular depth estimation from RGB is an ill-posed problem that can be solved only up to a geometric scale.
The combination of RGB and depth modalities form a rich source for mutual improvements where each sensor can benefit from the advantage of the other.\par
 Many pipelines have been proposed for a fusion of these two inputs~\cite{mal2018sparse,van2019sparse,qiu2019deeplidar,ma2019self,yang2019dense,xu2019depth}.
Ground truth annotations for this task, however, require elaborate techniques, manual adjustments and are subject to hardware noise or costly and time-consuming labeling.
The most prominent publicly available data for this task~\cite{uhrig2017sparsity} creates a ground truth by aligning consecutive raw LiDAR scans that are cleaned from measurement errors, occlusions, and motion artifacts in a post-processing step involving classical stereo reconstruction.
Even after the use of this additional data and tedious processing, the signal is not noise-free as discussed in~\cite{uhrig2017sparsity}.
To avoid such annotations, some methods perform self-supervision~\cite{ma2019self,yang2019dense,wong2020unsupervised}, where a photometric loss is employed with stereo or video data.
The dependence on additional data such as stereo or temporal sequences brings other problems such as line-of-sight issues and motion artifacts from incoherently moving objects.

Modern 3D engines are capable of rendering highly realistic virtual environments~\cite{ros2016synthia,Gaidon:Virtual:CVPR2016,dosovitskiy2017carla} with perfect ground truth. However, a significant domain gap between real and virtual scenes prevents from successful training on synthetic data only.\par
\noindent\textbf{Contributions and Outline.} In contrast to the self-supervised methods~\cite{ma2019self,yang2019dense,wong2020unsupervised}, we propose to use a domain adaptation approach to address the depth completion problem without real data ground truth as shown in Figure~\ref{fig:initial}. We train our method from the synthetic data generated with the driving simulator CARLA~\cite{dosovitskiy2017carla} and evaluate it on the real KITTI depth completion benchmark~\cite{uhrig2017sparsity}. The real LiDAR data is noisy with the main source of noise being the see-through artifacts that occur after projecting the LiDAR's point cloud to the RGB cameras. We propose an approach to simulate the see-through artifacts by generating data in CARLA multicamera set-up, employing random masks from the real LiDAR to sparsify the virtual LiDAR sensor, and projecting from the virtual LiDAR to the RGB reference frame. We further improve the model by filtering the noisy input in the real domain, thus obtaining a set of reliable points that are used as supervision. Finally, to reduce the domain gap between the RGB images, we use a CycleGAN~\cite{zhu2017unpaired} to transfer the image style from the real domain to the synthetic one.
%The LiDAR pattern is simulated in the synthetic data first by creating a depth map on the geometrically correct virtual LiDAR reference which is consecutively sparsified with random binary masks from the real domain.  We further propose a filtering method in the target domain to remove see-through points and obtain reliable LiDAR points to use as supervision.
%We further use semi-supervision and provide a pseudo dense depth map on the real domain that acts as a supervision signal using weight-averaged consistency targets to improve the network output in form of a student-teacher scheme~\cite{tarvainen2017mean}.
%This pulls our noisy model predictions during training towards correct estimates.
%With a detailed ablation study and analysis of the different network components, our method is 
%
We compare our approach to other state-of-the-art depth completion methods and provide a detailed analysis of the proposed components. The proposed domain adaptation for RGB-guided sparse-to-dense depth completion is a novel approach for the task of depth completion, which leads to significant improvements as demonstrated by the results.
%KM: repetition with below and and error in mm is our of context. We report in the test set of the KITTI depth completion benchmark an MAE of 280.42~mm and an RMSE of 1095.26~mm, which constitute the state-of-the-art amongst the methods that do not use ground  truth in real domain. 
To this end, our main contributions~are:

\begin{enumerate}
    \item A novel domain adaptation method for depth completion that includes geometric and data-driven sensor mimicking, noise filtering and image style adaptation. We demonstrate that adapting the synthetic sparse depth is crucial for improving the performance, whereas RGB adaptation is secondary.

    %\item The \textbf{first method} for guided depth completion trained with synthetic data and no real GT or additional sensors. 
    \item The improved  state-of-the-art results for the KITTI depth completion benchmark, amongst  ground truth free methods, by 6.4\% RMSE and 6.3\% MAE, and by 9.2\% RMSE and 10.4\% MAE when combining our pipeline with video self-supervision.

\end{enumerate}

\section{Related Work}
%\todo{Something about depth estimation, completion, supervision, domain adaptation}
%\ben{Reference more to what we do}
We first review related works on depth estimation using either RGB or LiDAR, and then discuss depth completion methods using Convolutional Neural Networks. 
\subsection{Unimodal Approaches}
\noindent\textbf{RGB Images.}
RGB based depth estimation has a long history~\cite{scharstein2002taxonomy,lazaros2008review,tippetts2016review} reaching from temporal Structure from Motion (SfM)~\cite{faugeras1988motion,huang2002motion} and SLAM~\cite{handa2014benchmark,mur2017orb,engel2018direct} to recent approaches that estimate depth from a static image~\cite{laina:2016:deeper,godard:2017:unsupervised,guo:2018:learning,li:2018:megadepth}.
Networks are either trained with full supervision~\cite{eigen:2014:depth,laina:2016:deeper} or use additional cameras to exploit photometric consistency during training~\cite{godard:2017:unsupervised,poggi:2018:learning}.
Some monocular depth estimators leverage a pre-computation stage with an SfM pipeline to provide supervision for both camera pose and depth~\cite{klodt2018supervising,yang2018deep} or incorporate hints from stereo algorthms~\cite{watson2019self}.
%The first work regarding deep learning based depth estimation comes from Eigen et al.~\cite{eigen:2014:depth}. They employ a CNN to predict a depth map from a static monocular image. Laina et al.~\cite{laina:2016:deeper} improve on this by leveraging a fully convolutional residual network with an efficient encoder-decoder structure.
%If multiple cameras are considered during training time, self-supervision with photo-consistency losses becomes possible by warping pixels from one view to the other given a depth estimate. In this way, not only do additional ground truth labels become unnecessary but also the calibration-based error propagation is diminished. Godard et al.~\cite{godard:2017:unsupervised} propose to use left-right stereo consistency for self-supervision while Poggi et al.~\cite{poggi:2018:learning} use trinocular supervision.
%If multiple cameras are considered during training time, self-supervision with photo-consistency losses becomes possible~\cite{godard:2017:unsupervised, poggi:2018:learning}, circumventing the need for ground truth labels and diminishing calibration-based errors. 
These approaches are in general tailored for a specific use case and suffer from domain shift errors, which has been addressed with stereo proxies~\cite{guo:2018:learning} or various publicly available pre-training sources~\cite{li:2018:megadepth}. The estimated depth often suffers from over-smoothing~\cite{godard2019digging} with wrongly inferred ``flying pixels'' in the free space close to depth discontinuities.\par
%, e.g. if one wants to transfer from indoor data to outdoor scenes or even change the camera design. 
%To address this drawback, Guo et al.~\cite{guo:2018:learning} use stereo matching as a proxy for monocular depth estimation to benefit from pre-training on synthetic data across domains and MegaDepth~\cite{li:2018:megadepth} trains using photos available online and multi-view stereo.

\noindent\textbf{Sparse Depth.}
While recent advantages in depth super-resolution~\cite{voynov2019perceptual,lutio2019guided} show good performance, they are not directly applicable to LiDAR data which is sparsely and irregularly distributed within the image.
Similar to super-resolution, a rectangular grid for the sampling was assumed in~\cite{riegler2016atgv}.
The sampling grid of the sparse depth signal is crucial for the depth completion task~\cite{uhrig2017sparsity}, which can be provided as a mask to the network, thus helping to densify the input.
While classical image processing techniques are used in in~\cite{ku2018defense},  an encoder-decoder architecture is applied for this task in~\cite{jaritz2018sparse}.
Other approaches~\cite{chodosh2018deep,eldesokey2019confidence} design more efficient architectures to improve the runtime performance.
%Chodosh et al.~\cite{chodosh2018deep} design an efficient architecture leveraging dictionary learning for this task and Eldesokey et al.~\cite{eldesokey2019confidence} significantly reduce the amount of networks parameters usually needed.
\subsection{Depth Completion from RGB and LiDAR}
Most recent solutions to depth completion leverage deep neural networks. These can be divided into (a) supervised and (b) self-supervised approaches.\par
%In addition, there are also learning-free methods, i.e., classical methods, described next \ben{are there some? And where are they described}
\noindent\textbf{Supervision and Ground Truth.}
Usually, an encoder-decoder network is used to encode the different input signals into a common latent space where feature fusion is possible and a decoder reconstructs an output depth map~\cite{van2019sparse,qiu2019deeplidar,lee2019depth,mal2018sparse,xu2019depth}.
Different additional random sampling strategies can increase the density of the input signal~\cite{mal2018sparse}  while fusing 2D and 3D representations~\cite{Chen_2019_ICCV} can
improve depth boundaries. 
The noise problem has been targeted with local and global information in~\cite{van2019sparse}.
Other methods~\cite{qiu2019deeplidar,xu2019depth} leverage different input modalities such as surface normals to increase the amount of diversity in the input data. The publicly available dataset KITTI~\cite{geiger2012we,uhrig2017sparsity} includes real driving scenes where a stereo RGB camera system is fixed on the roof of a car along with a LiDAR scanner that acquires data while the car is driving.
A post-processing stage fuses several LiDAR scans and filters outliers with the help of stereo vision to provide labeled ground truth. While this process is intricate and time-consuming, further error is accumulated from calibration and alignment~\cite{uhrig2017sparsity}.\par
\noindent\textbf{Self-Supervised Approaches.}
Another view either from a second camera or a video sequence can be used for self-supervision.
Temporal information and mutually predicted poses between RGB frames were used in~\cite{ma2019self} for self-supervision with a photometric loss on the reprojected image. A probabilistic formulation was proposed in~\cite{yang2019dense} with a conditional prior within a MAP estimation, which also leverages stereo information. A non-learning method was used in \cite{wong2020unsupervised} to form a spatially dense but coarse depth approximation from the sparse points, where the coarse approximation was then refined using another network. A photometric loss was also used in \cite{wong2020unsupervised}, where a separate network predicted the poses between RGB frames obtained from a video sequence.\par
% Other scholars~\cite{wong2019voiced} leverage additional modalities such as IMU for pose prediction to also perform an image reprojection \adrian{Not sure this one uses IMUs}.\par
\noindent\textbf{Synthetic Data.}
For monocular depth estimation, two domain adaptation approaches used style-transfer methods~\cite{atapour2018real,zheng2018t2net}.  Sparse-to-dense methods, however, have used synthetic data without any adaptation \cite{atapour2019complete,yang2019dense,qiu2019deeplidar} so far. Training on synthetic data requires a high rendering quality~\cite{mayer2018makes}. To this end, synthesizing driving scenarios has also been researched: SYNTHIA~\cite{ros2016synthia} provides synthetic urban images together with semantic annotations, while Virtual KITTI~\cite{Gaidon:Virtual:CVPR2016} and its point cloud variant~\cite{3dsemseg_ICCVW17} constitutes a dataset with synthetic renderings and point clouds that closely match the videos of the KITTI dataset~\cite{geiger2012we} including pixel-perfect semantic ground truth and depth labels. The CARLA simulator~\cite{dosovitskiy2017carla} was introduced allowing for photo-realistic simulations of driving scenarios, which we utilize to generate realistic RGB images. A LiDAR simulator using ray-casting and a learning process to drop points was proposed in \cite{manivasagam2020lidarsim}, which was tested in detection and segmentation tasks, but is not publicly available. While LiDAR scans can also be simulated with CARLA via ray-casting, the car shapes are approximated with cuboids losing much detail. We leverage the simulator z-buffer to estimate fine-granular depth and then sparsify the signal to simulate LiDAR scans, thus closely matching the real domain. %km: this is better for intro although being the first is not enough.
%To the best of our knowledge, 

%Domain Adaptation
%    ? Modulating Inputs
%    ? Modulating Features / Embeddings
%    ? Minimal GT on real
%    History:
%        known labels, supervised learning
%        training on unlabelled examples: add noise to clean prediction and pull the noisy prediciton to the clean one
%        with one prediction (DSS, ladder gamma model)
%        two predictions (PEA, PI model, see Tarvainen paper)
%        use ensamble many models that form a teacher prediction
%        a teacher prediction pulls a noisy student prediction (temporal ensambling, mean teacher)
%        correct noisy sample with adversarial method and pull towards prediction (virtual adv. training) or pull towards closest class (entropy minimization)
%   Semi-supervised approaches

\begin{figure*}[t]
    \centering
    \includegraphics[width=\textwidth,clip]{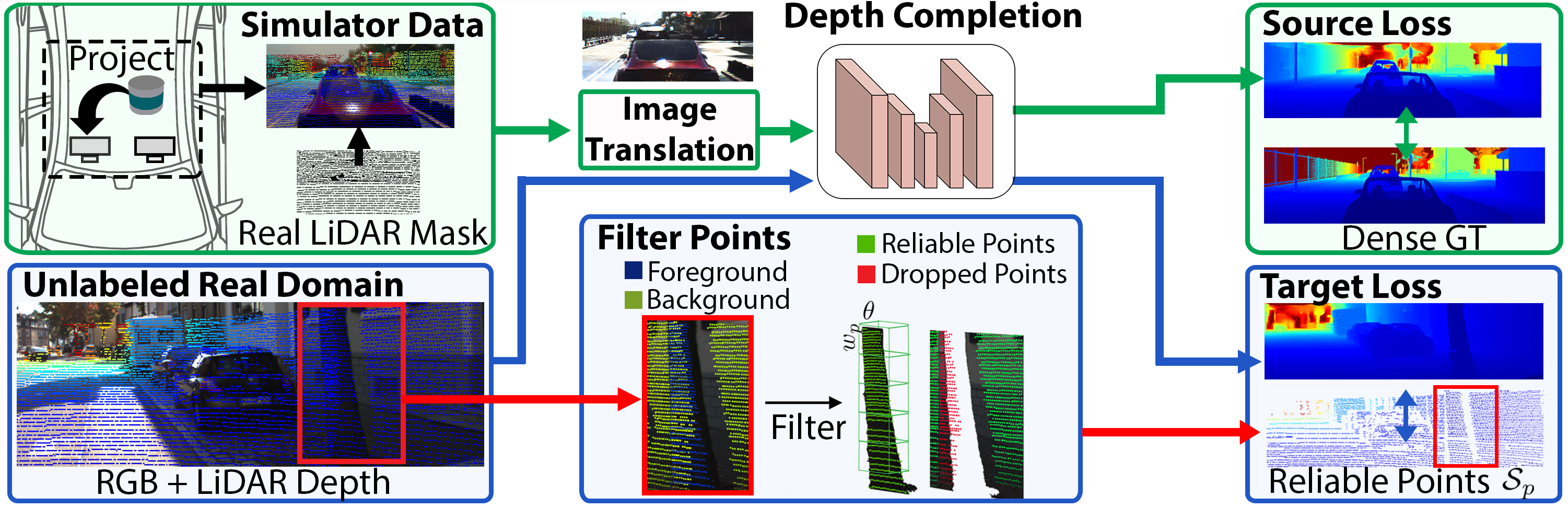}
    \caption{The main components of our method. We use a simulator with a multicamera set-up and real LiDAR binary masks to transform the synthetic dense depth map into a noisy and sparse depth map. We train in a two-step manner: the green blocks are used in the 1st and the 2nd step of training while the blue blocks in the 2nd step only. In the filtering block, green points are the reliable points $\mathcal{S}_p$ and red points are dropped. The \textit{Image Translation} network is pretrained using a CycleGAN approach~\cite{zhu2017unpaired}}
    \label{fig:method}
\end{figure*}
\section{Method}
\label{sec:method}
%We now present our proposed method consisting on two sub-blocks: source data adaptation and retrieval of reliable supervision from the target noisy LiDAR information.
Our method, shown in Figure~\ref{fig:method}, consists of two main components that include an adaptation of the synthetic data to make it similar to the real data, as well as a retrieval of reliable supervision from the real but noisy LiDAR signal.
\subsection{Data Generation via Projections}
\label{sec:projections}
Supervised depth completion methods strongly rely on the sparse depth input, achieving good performance without RGB information~\cite{ma2019self,van2019sparse}. 
To train a completion model from synthetic data that works well in the real domain we need to generate a synthetic sparse input that reflects the real domain distribution. Instead of simulating a LiDAR via ray-casting, which is computationally expensive and hard to implement~\cite{yue2018lidar}, we leverage the z-buffer of our synthetic rendering engine to provide a dense depth ground truth at first.\par
Previous approaches used synthetic sparse data to evaluate a model in indoor scenes or synthetic outdoor scenes~\cite{huang2018hms,jaritz2018sparse,yang2019dense}. To sparsify the data a Bernoulli distribution per pixel is used in some works~\cite{huang2018hms,jaritz2018sparse,mal2018sparse} which, given a probability $p_B$ and a dense depth image $x_{D}$, samples each of the pixels $x_{D,k}$ by either keeping the value $x_{D,k}$ with probability $p_B$ or setting its value to $0$ with probability $(1-p_B)$, thus generating the sparse depth $x^{s_B}_{D}$.
We argue that using $x^{s_B}_{D}$ does not simulate well a real LiDAR input, thus a model trained with $x^{s_B}_{D}$ does not perform well in the real domain. Our results in Section~\ref{sec:experiments} support this observation. There are two reasons for the drop of performance in the real LiDAR data. Firstly, the distribution of the points $x^{s_B}_{D}$ does not follow the LiDAR sparse distribution. Secondly, there is no noise in the sampled points, as we directly sample from the ground truth. We now propose an approach to address these two issues.\par
\noindent\textbf{Mimicking LiDAR Sampling Distribution.} To simulate a pattern similar to a real LiDAR, we propose to sample at random the real LiDAR inputs $x^{s}_{R, D}$ from the real domain similarly to \cite{atapour2019complete}. We use $x^{s}_{R, D}$ to generate a binary mask $M_{L}$, which is 1 in $M_{L,k}$ if $x^{s}_{R, D, k}>0$ and 0 if $x^{s}_{R, D, k}=0$. We then apply the masks to the dense synthetic depth data by $x^{s_M}_D = M_L\odot x_D$. This approach adapts the synthetic data directly to the sparsity level in the real domain without the need to tune it depending on the LiDAR used.\par
%An alternative approach would be sampling the image using the LiDAR specifications, including the number of lasers and angular resolution, however we would need to add jitter and points dropping to mimic real LiDAR data adding extra parameters to tune.\par
% Another approach is sampling the image using the LiDAR specifications, however we would need to add jitter to mimic real LiDAR data adding extra parameters to tune.\par

\begin{figure}[t]
    \begin{subfigure}[t]{0.5\textwidth}
        \centering
        \includegraphics[trim={0cm 0cm 0 4cm},width=\linewidth,clip]{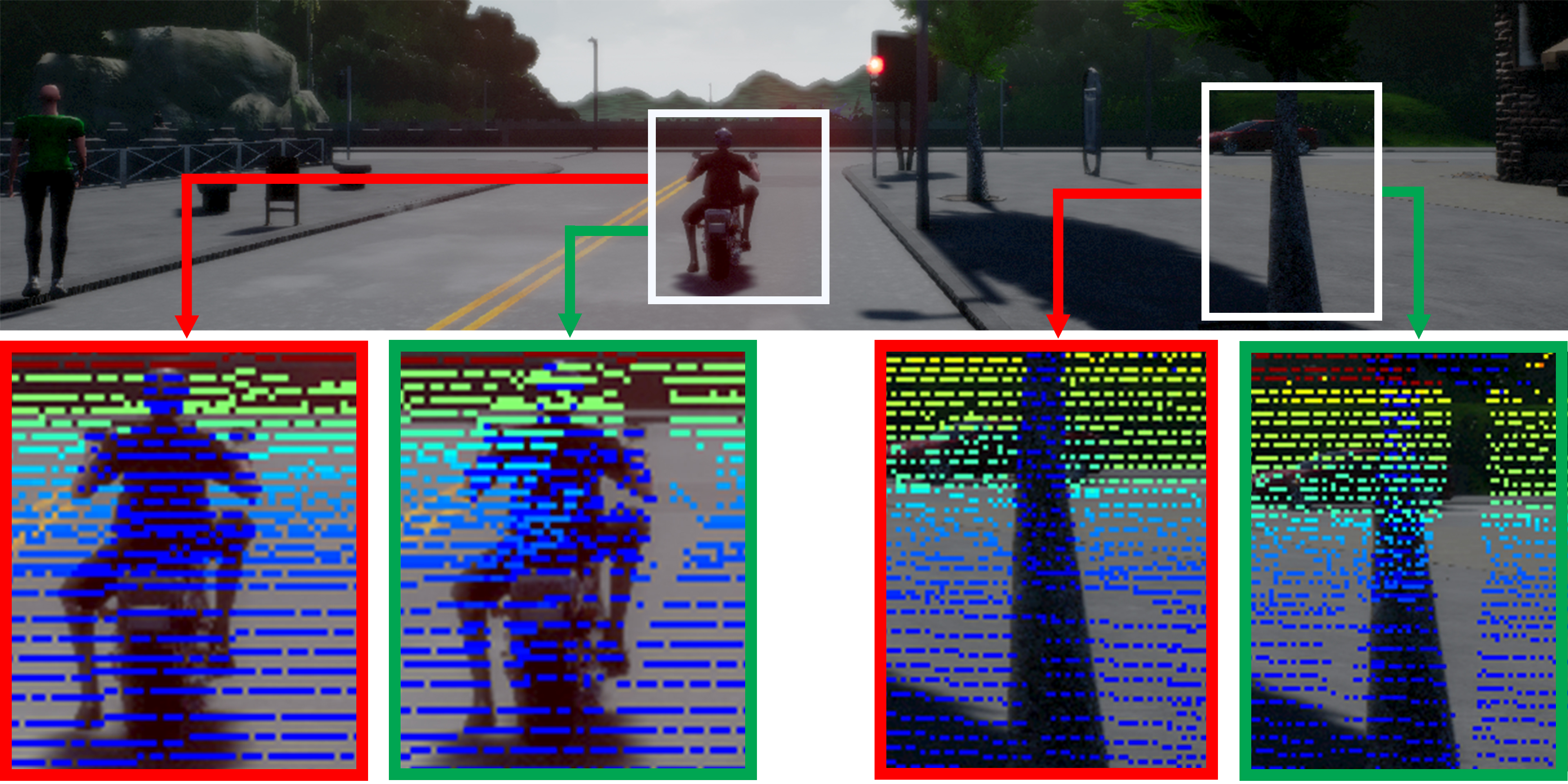}
    \end{subfigure}%
    ~ 
    \begin{subfigure}[t]{0.5\textwidth}
        \centering
        \includegraphics[trim={0cm 0cm 0 0}, width=\linewidth,clip]{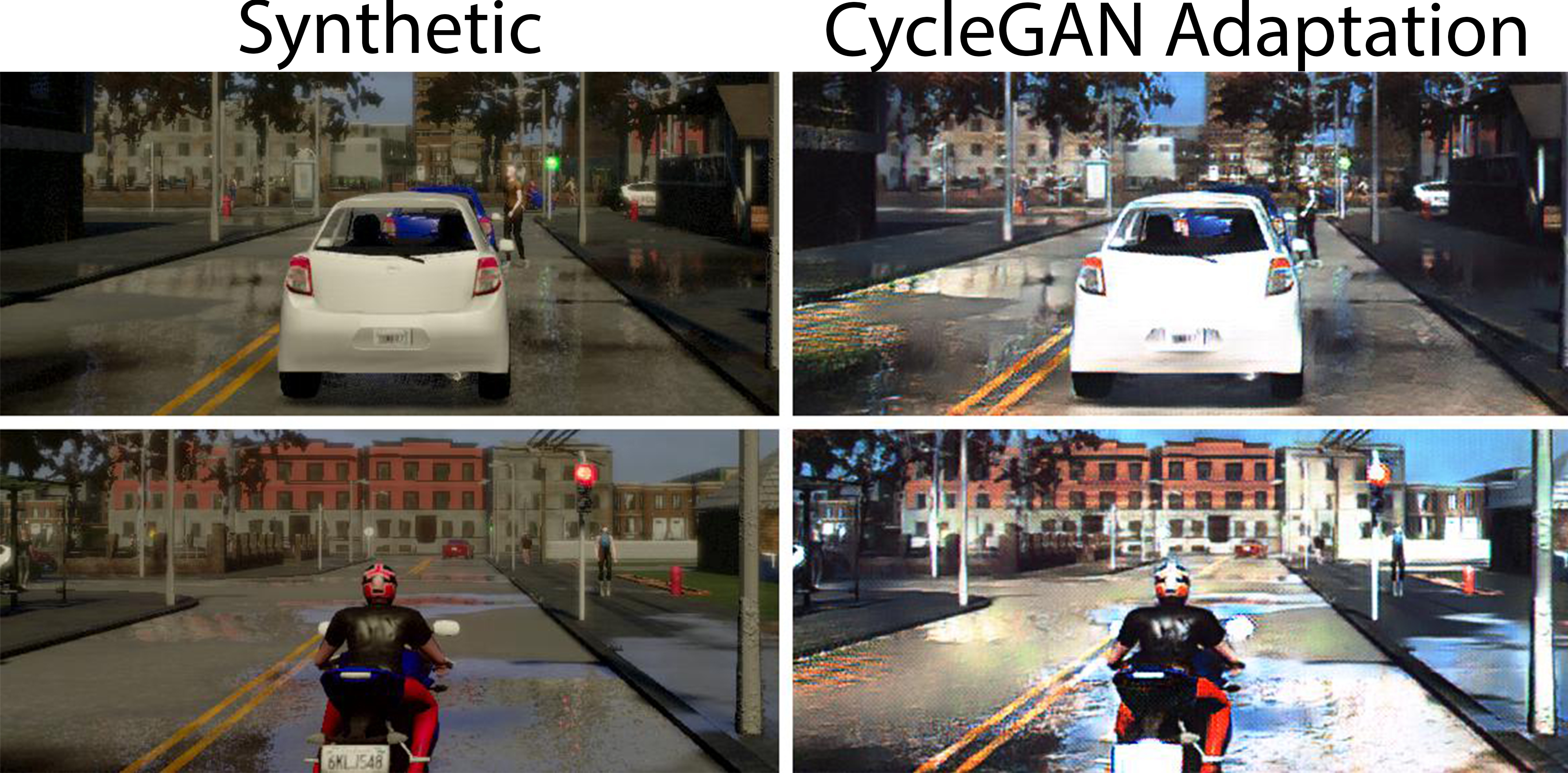}
    \end{subfigure}
    \caption{\textit{Left:} Example of the generated projection artifacts in the simulator.
    The zoomed-in areas marked with red rectangles correspond to $x^{s_M}_D$ and the zoomed-in areas marked with green rectangles to $x^{s_P}_D$, where we can see simulated projection artifacts, e.g. see-through points on the left side of the motorcyclist. \textit{Right:}~We reduce the domain gap in the RGB modality using a CycleGAN approach. We show synthetic CARLA images and the resulting adapted images}
    \label{fig:points_projection}
\end{figure}
\noindent\textbf{Generating Projection Artifacts.} Previous works use noise-free sparse data to pre-train~\cite{qiu2019deeplidar} or evaluate a model~\cite{jaritz2018sparse} with synthetic data. However, simulating the noise of real sparse data can reduce the domain gap and improve the adaptation result. Real LiDAR depth contains noise from several sources including the asynchronous acquisition due to the rotation of lasers, dropping of points due to low surface reflectance and projection errors. Simulating a LiDAR sampling process by modelling all of these noise sources can be costly and technically difficult as a physics-based rendering engine with additional material properties is necessary to simulate the photon reflections individually. We propose a more pragmatic solution and use the z-buffer of a simulator by assuming that the dominating noise is a consequence of the point cloud projection to the RGB camera reference frame. For such a simulation, the error becomes twofold. Firstly, the 3D points are not exactly projected on the pixel center which produces a minor quantization error. Secondly, as we are projecting a sparse point cloud arising from another viewpoint, we do not have a way to filter the overlapping points by depth. This creates the see-through patterns that do not respect occlusions as shown in Figure~\ref{fig:points_projection} which is also observed in the real domain~\cite{cheng2019noise}. Therefore, a simple point drawing from a depth map at the RGB reference cannot recreate this effect and such method does not perform well in the real domain.
%We propose a solution that uses the standard z-buffer of a simulator by assuming that the noise that dominates the input is a consequence of projecting the point cloud information to another camera.

%In setups with a depth sensor, such as LiDAR, and RGB images, the LiDAR points are usually projected to the RGB images generating two source of errors.

%Second, and a bigger source of error, as we are projecting a sparse depth we do not have any method to filter the overlapping points by depth, creating the see-through pattern shown in Figure~\ref{fig:points_projection}.

%The reason for the non-optimal performance is that we do not have any noise in the input depth points and the model learns to not correct them.\ben{see above. Maybe rephrasing.}\par
To recreate this pattern, we use the CARLA simulator~\cite{dosovitskiy2017carla}, which allows us to capture multicamera synchronized synthetic data.
Our CARLA set-up mimics the camera distances in KITTI~\cite{geiger2012we}, as our benchmark is the KITTI depth completion dataset~\cite{uhrig2017sparsity}. Instead of a LiDAR, we use a virtual depth camera. The set-up is illustrated in Figure~\ref{fig:initial}.
%We place the LiDAR camera, which is a standard dense depth camera using the simulator z-buffer, at a distance from the RGB cameras.
As the data is synthetic, the intrinsic and extrinsic parameters needed for the projections are known.
After obtaining the depth from the virtual LiDAR camera, we sparsify it using the LiDAR masks resulting in $x^{s_M}_D$, which is then projected onto the RGB reference with
\begin{equation}
    x^{s_P}_D = K_{RGB} P^{L}_{RGB} K_{L}^{-1} x^{s_M}_D
\label{eq:lidar_mask_data}
\end{equation}
where $K_{L}$, $K_{RGB}$ are the LiDAR and RGB camera intrinsics and $P^{L}_{RGB}$ is the rigid transformation between the LiDAR and RGB reference frame.
The resulting $x^{s_P}_D$ is the projected sparse input to either left or right camera. %Figure~\ref{fig:points_projection} shows the simulated see-through artifacts.\par

\subsection{RGB Adaptation} 
Similarly to domain adaptation for depth estimation methods~\cite{zheng2018t2net,atapour2018real,zhao2019geometry}, we address the domain gap in the RGB modality with style translation from synthetic to real images. Due to the added computational complexity of adapting high-resolution images, we first train a model to translate from synthetic to real using a CycleGAN \cite{zhu2017unpaired} approach. The generator is not further trained and is used to translate the synthetic images to the style of real images, thus reducing the domain gap as shown in Figure~\ref{fig:points_projection}.\par 
 %To do so we use a classical edge detector in both the source and target domain, so we embed them in a common space where the domain gap is decreased. 
\subsection{Filtering Projection Artifacts for Supervision}
\label{sec:filt}
%Past domain adaptation research showed that using pseudo-labels in the target domain increases the performance in several settings~\cite{tang2012shifting, zou2018unsupervised}. Similarly, we aim to give a supervision in the target domain. 
In a depth completion setting, the sparse depth input can be used as supervision data, similarly to~\cite{ma2019self}. However, the approach from \cite{ma2019self} did not take into account the noise present in the data. The given LiDAR input is precise in most points with an error of only a few centimeters. However, due to the noise present, some points cannot be used for supervision, such as the see-through points, which have errors in the order of meters. Another method~\cite{cheng2019noise} also used the sparse input as guidance for LiDAR-stereo fusion while filtering the noisy points using stereo information. We propose to filter the noisy input without using additional sensor data such as a stereo pair as this may not always be available.\par
%However, as we have shown, the input in some points is noisy, with errors of meters, and training with these inputs can either limit the achievable performance or even hurt the performance of the approach.\par
Our goal is to find a set of reliable sparse points $\mathcal{S}_p$, likely to be correct, for supervision based on the assumption used in Section~\ref{sec:projections}, i.e., the main source of error are the see-through points after projection. We assume that in any given local window there are two modes of depth distribution, approximated by a closer and a further plane.  
We show an overview of the idea in Figure~\ref{fig:method}. The points from the closer plane are more likely to be correct as part of the occluding objects. To retrieve $\mathcal{S}_p$ we apply a minimum pooling with window size $w_p$ yielding a minimum depth value $d_m$ per window. Then, we include  in  $\mathcal{S}_p$ the points  $s\in[d_m, d_m+\theta]$ where $\theta$ is a local thickness parameter of an object. The number of noisy points not filtered out depends on the window $w_p$ and object thickness $\theta$, e.g. larger windows remove more points but the remaining points are more reliable. We use the noise rate $\eta$, which is the fraction of noisy points as introduced in noisy labels literature~\cite{li2019learning,han2018co,zhang2018generalized}, to select $w_p$ and  $\theta$ in the synthetic validation set, thus not requiring any ground truth in the real domain.\par
After the filtering step, a certain number of false positives remains. The noisy points in $\mathcal{S}_p$ are more likely to be further away from the dense depth prediction $\hat{y}$, hence BerHu will give more weight to those outliers. To provide extra robustness against these false positives we use in the real domain a Mean Absolute Error (MAE) loss instead of the Reverse Huber (BerHu) loss used in the synthetic domain, as MAE weights all values equally, showing more robustness to the noise.\par

\subsection{Summary of Losses}
\noindent Our proposed loss is
\begin{equation}
    \mathcal{L}=\lambda_S\mathcal{L}_S + \lambda_R\mathcal{L}_R
\end{equation}
where $\mathcal{L}_S$ is the loss used for the synthetic data, $\mathcal{L}_R$ the loss used for the real data and $\lambda_S$ and $\lambda_R$ are hyperparameters.\par
% \km{shouldn't that be $\lambda$ and $1-\lambda$?}
We use a two-step training approach similar to past domain adaptation works using pseudo-labels~\cite{zou2018unsupervised,tang2012shifting}, aiming first for good performance in the synthetic data before introducing noise in the labels. First, we set $\lambda_S=1.0$ and $\lambda_R=0.0$, to train only from the synthetic data. For $\mathcal{L}_S$ we use a Reverse Huber loss, which works well for depth estimation problems~\cite{laina:2016:deeper}. Hence, we define $\mathcal{L}_S$ as
 \begin{equation}
     \mathcal{L}_S = \frac{1}{b_S}\sum_i \frac{1}{n_{i}} \sum_k \mathcal{L}_{bh}(\hat{y_k}, y_k)
\end{equation}
where $b_S$ is the synthetic batch size, $n_i$ the number of ground truth points in image $i$, $\hat{y}$ is the predicted dense depth, $y$ is the ground truth depth and $\mathcal{L}_{bh}$ is the Reverse Huber loss~\cite{zwald2012berhu}.\par
% the Reverse Huber loss $\mathcal{L}_{bh}$~\cite{zwald2012berhu} is 
% \begin{equation}\label{eq:berhu}    
%      \mathcal{L}_{bh}(\hat{y}_k, y_k)=\begin{cases} \frac{(\hat{y}_k - y_k)^2 + c^2}{2c} & \mbox{if } |\hat{y}_k - y_k| > c  \\ |\hat{y}_k - y_k|  &  \mbox{if } |\hat{y}_k - y_k|\leq c  \end{cases}
% \end{equation}
% where $c=\delta\cdot\max_k{|\hat{y}_k - y_k|}$ and we set $\delta=0.05$, which is the default value in the official code of the model we use~\cite{van2019sparse}.\par
In the second step we set $\lambda_S=1.0$ and $\lambda_R=1.0$ as we introduce  real domain data into the training process using $\mathcal{S}_p$ for supervision. We define $\mathcal{L}_R$ as
 \begin{equation}
     \mathcal{L}_R = \frac{1}{b_R}\sum_i\frac{1}{\#(\mathcal{S}_{p,i})}\sum_k|\hat{y}_k - y_k|
 \end{equation}
 where $b_R$ is the real domain batch size and $\#(\mathcal{S}_{p,i})$ is the cardinality of the set of reliable points $\mathcal{S}_{p}$ for an image $i$.\par

\begin{table}[t]
\vspace{-2mm}
\caption{Ablation study on the selected validation set. \textit{Bernoulli} refers to training using $x^{s_B}_D$, \textit{Mask} to training using $x^{s_M}_D$, \textit{Proj.} to training using $x^{s_P}_D$, \textit{CycleGAN RGB} to translating the synthetic RGB images to the style of the real domain, and \textit{BerHu} refers to using BerHu for real data supervision. All of \textit{2nd Step} results use \textit{LiDAR Mask + Proj + CycleGAN RGB}}
\label{tab:ablation_study}
\begin{center}
\scriptsize
\begin{tabular}{lcccc}\toprule
        {Model} & {RMSE} & {MAE} & {iRMSE} & {iMAE} \\
    \midrule
    
        \multicolumn{1}{l}{\textbf{1st Step: Only Synthetic Supervision}} \\
        \text{\textit{Syn. Baseline 1:} Bernoulli ($p_B$=0.1)}  & 1975.14 & 458.41 & 8.21 & 2.25 \\
        \text{\hspace{1em}+ Proj.} & 3286.73 & 1253.42 & 14.58 & 6.92 \\
        \text{\textit{Syn. Baseline 2:} LiDAR Mask} & 1608.32 & 386.49 & 7.13 & 1.76\\
        \text{\hspace{1em}+ Proj.} & 1335.00 & 342.16 &  5.41 & 1.55\\
%        \text{\hspace{1em}+ Proj. + No RGB} & 1196.90 & 305.43 & 4.06 & 1.32\\
        \text{\hspace{1em}+ Proj. + CycleGAN RGB} & 1247.53 & 308.08 & 4.54 & 1.34\\
    \midrule
    \multicolumn{1}{l}{\textbf{2nd Step: Adding Real Data}} \\
        \text{No Filter} & 1315.74 & 315.40 & 4.70 & 1.40\\
        \text{$\mathcal{S}_p$+BerHu} & 1328.76 & 320.23 & 4.25 & 1.33\\
        \text{\textit{Full Pipeline:} $\mathcal{S}_p$} &\textbf{ 1150.27} & \textbf{281.94} & \textbf{3.84} & \textbf{1.20}\\
    \midrule    
    \midrule
    \text{Real GT Supervision} & 802.49 & 214.04 & 2.24 & 0.91 \\
    \bottomrule
\end{tabular}
\end{center}
\vspace{-1em}
\end{table}
\section{Experiments}\label{sec:experiments}
\noindent We use PyTorch 1.3.1~\cite{paszke2017automatic} and an NVIDIA 1080 Ti GPU as well as the official implementation of FusionNet~\cite{van2019sparse} as our sparse-to-dense architecture. The batch size is set to 4 and we use Adam~\cite{kingma2015adam} with a learning rate of 0.001. For the synthetic data, we train using $x^{s_P}_D$ by randomly projecting to the left or right camera with the same probability. In the first step of training, we use only synthetic data (i.e., $\lambda_S=1.0$, $\lambda_R=0.0$, $b_S=4$ and $b_R=0$) until performance plateaus in the synthetic validation set. In the second step, we mix real and synthetic images setting $\lambda_S=1.0$, $\lambda_R=1.0$, $b_S=2$, $b_R=2$, $w_p=16$~pixels and $\theta=0.5$~m, and train for 40,000 iterations.\par
To test our approach, data from a real LiDAR+RGB set-up is needed as we address the artifacts arising from projecting the LiDAR to the RGB camera. There are no standard real LiDAR+RGB indoor depth completion datasets available. In NYUv2 \cite{silberman2012indoor} the dense ground-truth is synthetically sparsified using Bernoulli sampling, while VOID \cite{wong2020unsupervised} provides sparse depth from visual inertial odometry that contains no projection artifacts. Thus, the KITTI depth completion benchmark~\cite{uhrig2017sparsity} is our real domain dataset, as it provides paired real noisy LiDAR depth with RGB images, along with denser depth ground truth for testing. We evaluate our method in the selected validation set and test set, each containing 1,000 images. Following~\cite{van2019sparse}, we train using images of 1216x256 by cropping their top part. We evaluate on the full resolution images of 1216x356. The metrics used are Root Mean Squared Error (RMSE) and Mean Absolute Error (MAE), reported in mm, and inverse RMSE (iRMSE) and inverse MAE (iMAE), in 1/km.\par
% The metrics used for evaluation are the Root Mean Squared Error of the depth (RMSE) and the inverse depth (iRMSE) and the Mean Absolute Error of the depth (MAE) and the inverse depth (iMAE).
\noindent\textbf{Synthetic Data.} We employ CARLA 0.84~\cite{dosovitskiy2017carla} to generate synthetic data using the camera set-up in Figure~\ref{fig:initial}. We collect images from 154 episodes resulting in 18,022 multicamera images for training and 3,800 for validation. An episode is defined as an expert agent placed at random in the map and driving around while collecting left and right depth+RGB images, as well as the virtual LiDAR depth. We use for the virtual LiDAR camera a regular dense depth camera instead of the provided LiDAR sensor in CARLA because the objects in the LiDAR view are simplified (e.g., CARLA approximates the cars using cuboids). The resolution of the images is 1392x1392 with a Field Of View of 90\degree. To match the view and image resolution in KITTI, we first crop the center 1216x356 of the image and then the upper part of 1216x256. To adapt the synthetic RGB images, we train the original implementation of CycleGAN \cite{zhu2017unpaired} for 180,000 iterations.\par

\begin{table}[t]
\vspace{-2mm}
\caption{Results in the selected validation set depending on the input type. Three different runs are averaged to reduce variability}
\label{tab:input_type}
\begin{center}
\scriptsize
\begin{tabular}{lcccc} \toprule
        {Input Data} & {RMSE} & {MAE} & {iRMSE} & {iMAE} \\
    \midrule
        \text{Only Sparse Depth} & 1175.54 & 290.51 & 4.11 & 1.27  \\
        \text{+ RGB} & 1167.83 & 289.86 & 3.87 & 1.28\\
        \text{+ Img. Transfer from \cite{zheng2018t2net}} & 1184.39 & 306.66 & 3.99 & 1.36  \\

        \text{+ CycleGAN RGB} & \textbf{1150.27} & \textbf{281.94} & \textbf{3.84} & \textbf{1.20}\\
    \bottomrule
\end{tabular}
\end{center}

\end{table}

% \begin{table}[h]
% \begin{center}
% \scriptsize
% \begin{tabular}{lcccc} \toprule
%     {Model} & {RMSE} & {MAE} & {iRMSE} & {iMAE} \\
%     \midrule
%     \textbf{Unsupervised} & & & & \\
%     \text{DDP~\cite{yang2019dense}} & 1285.14 & 353.16 & 3.69 & 1.37\\
%     \midrule
%     \textbf{Self-Supervised}  & & & &\\
%         \text{SS-S2D~\cite{ma2019self}} & 1299.85 & 350.32 & 4.07 & 1.57\\
%     \text{DDP+Stereo~\cite{yang2019dense}} & 1263.19 & 343.46 & 3.58 & 1.32\\

%     \midrule
%     \textbf{Domain Adap.}  & & & &\\
%     \text{Ours} & \textbf{1093.55} & \textbf{274.49} & \textbf{3.41} & \textbf{1.16} \\
%     \midrule
%     \midrule
%     \textbf{Supervised} & & & &\\
%     \text{DDP~\cite{yang2019dense}} & 836.00 & 205.40 & 2.12 & 0.86 \\
%     \text{S2D~\cite{ma2019self}} & 814.73 & 249.95 & 2.80 & 1.21\\
%     \text{FusionNet~\cite{van2019sparse}} & 772.87 & 215.02 & 2.19 & 0.93\\
%     \text{DeepLiDAR~\cite{qiu2019deeplidar}} & 758.38 & 226.50 & 2.56 & 1.15\\
%     \bottomrule
% \end{tabular}
% \end{center}
% \caption{Comparison of results in the official KITTI test set obtained through the online evaluation tool of~\cite{uhrig2017sparsity}. MAE \& RMSE are expressed in mm and iRMSE \& iMAE in 1/km.}
% \label{tab:test_comp}
% \end{table}

\subsection{Ablation Study}\noindent
We include an ablation study in Table~\ref{tab:ablation_study} using the validation set. For the result of the whole pipeline, we average the results of three different runs to account for training variability. All of the proposed modules provide an increase in accuracy. \par
\begin{figure*}[t]
    \begin{center}
    \includegraphics[trim={0cm 0cm 0 0},width=\textwidth,clip]{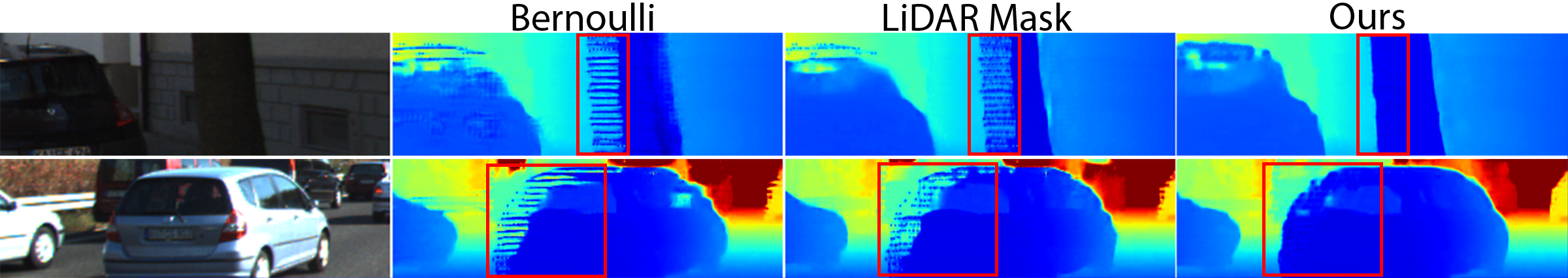}
    \end{center}
    \vspace{-1em}
    \caption{Qualitative results with different training methodologies. \textit{Bernoulli} refers to training using $x^{s_B}_D$, \textit{LiDAR Mask} to training using $x^{s_M}_D$ and \textit{Ours} to our full pipeline. Both rows show projection artifacts which we deal with correctly}
    \label{fig:qualitative_val}
\end{figure*}

\noindent\textbf{CARLA Adaptation.} Table~\ref{tab:ablation_study} shows that reprojecting the sparse depth is as important as matching the LiDAR sampling pattern, decreasing the RMSE by 32.4\%  when used jointly, i.e training with $x^{s_P}_D$ instead of $x^{s_B}_D$. Table~\ref{tab:ablation_study} also shows that training with the reprojected $x^{s_B}_D$ results in worse performance compared to training with $x^{s_B}_D$, showing that it is the combination of using a LiDAR distribution of points and projection to another camera which reduces the domain gap. Even though CycleGAN mostly adapts the brightness, contrast and colors of the images as shown in Figure \ref{fig:points_projection}, using CycleGAN adaptation to the style of the real domain further reduces the RMSE by 6.6\% when training with $x^{s_P}_D$ examples. 
%\todo{When training only with synthetic data, i.e. \textit{1st Step} in Table \ref{tab:ablation_study}, only using the sparse depth as input, \textit{+ No RGB}, performs better than using CycleGAN examples, which points towards a larger domain gap in the RGB that hinders the performance. Table \ref{tab:input_type} shows that for the full pipeline, i.e. after the 2nd Step, using only sparse depth as input performs worse, as will be discussed later in this section}. 
Figure~\ref{fig:qualitative_val} includes some predictions when training using $x^{s_B}_D$, $x^{s_M}_D$ and $x^{s_P}_D$ for examples with projection artifacts, showing that training using $x^{s_P}_D$ in the synthetic images is crucial to deal with the noisy input in the real domain.\par
\noindent\textbf{Introducing Real Domain Data.} Introducing the reliable points $\mathcal{S}_p$ as supervision in the real domain alongside the MAE loss function increases the performance as Table~\ref{tab:ablation_study} shows. If we use BerHu along with $\mathcal{S}_p$ supervision, the method deteriorates as the noisy points are likely to dominate the loss even if the noise rate $\eta$ in $\mathcal{S}_p$ is low. Using MAE without filtering also drops the performance, in this case due to the high noise rate $\eta$. These results show that using the noisy LiDAR points for supervision as in~\cite{ma2019self,wong2020unsupervised} is detrimental to the depth completion performance. To understand the noise level in the data, we define a point to be noisy if it is at least 0.3~meters away from the ground truth. Using this metric, the noise rate~$\eta$ for the points with available ground truth is 5.8\%, with our filtering method decreasing $\eta$ to 1.7\% while dropping 45.8\% of input points. The results suggest that the value of $\eta$ in $\mathcal{S}_p$ is more important than the total amount of points used for supervision. Our method shows a clear improvement compared to both synthetic baselines (\textit{Syn. Baseline}) defined in Table~\ref{tab:ablation_study}.\par
%These results show that the noisy LiDAR points cannot be directly used for supervision as was done in~\cite{ma2019self}, needing to filter them first.
\begin{table}[t]
\vspace{-2mm}
\caption{Comparison of results in the KITTI selected validation set and the official online test set. Non-learning methods results are from~\cite{qiu2019deeplidar}. \textit{DA Base} is our Domain Adaptation baseline formed by CycleGAN \cite{zhu2017unpaired} + LiDAR Masks.}
\label{tab:validation_comp}
\begin{center}
\scriptsize
\begin{tabular}{lcccccccccc} \toprule
     & & \multicolumn{4}{c}{\textbf{Validation Set}} & & \multicolumn{4}{c}{\textbf{Online Test Set}}\\
     \cmidrule(lr){3-6}\cmidrule(lr){8-11}
    {Model} & Param. & {RMSE} & {MAE} & {iRMSE} & {iMAE} &\ \ \ & {RMSE} & {MAE} & {iRMSE} & {iMAE}\\
    \midrule
    \textbf{Non-learning} & & & & \\
    \text{Cross-Bilateral~\cite{silberman2012indoor}} & - & 2989.02 & 1200.56 & 9.67 & 5.08 & & - & - & - & -\\
    \text{Fast Bilateral Sol.~\cite{barron2016fast}} & - & 3548.87 & 1767.80 & 26.48 & 9.13 & & - & - & - & -\\
    \text{TGV~\cite{ferstl2013image}} & - & 2761.29 & 1068.69 & 15.02 & 6.28 & & - & - & - & -\\
    \midrule
    \textbf{Unsupervised}  & & & &\\
    \text{DDP~\cite{yang2019dense}} & 18.8M & 1325.79 & 355.86 & - & - & & 1285.14 & 353.16 & 3.69 & 1.37\\
    \midrule
    \textbf{Self-Supervised}  & & & &\\
    \text{SS-S2D~\cite{ma2019self}} & 27.8M & 1384.85 & 358.92 & 4.32 & 1.60 & & 1299.85 & 350.32 & 4.07 & 1.57\\
    \text{DDP+Stereo~\cite{yang2019dense}} & 18.8M & 1310.03 & 347.17 & - & - & & 1263.19 & 343.46 & 3.58 & 1.32\\
    \text{VOICED~\cite{wong2020unsupervised}} & 9.7M & 1239.06 & 305.06 & 3.71 & 1.21 & & 1169.97 & 299.41 & 3.56 & 1.20 \\
    \midrule
    \textbf{Domain Adap.}  & & & &\\
        \text{DA Base} & 2.6M & 1630.31 & 423.70  & 6.64 & 1.98 & & - & - & - & -\\
         \text{\hspace{1em} + Disc. Out. \cite{pilzer2018unsupervised}} & 2.6M & 1636.89  & 390.59  & 6.78 & 1.78 & & - & - & - & -\\
        \text{\hspace{1em} + Disc. Feat. \cite{zheng2018t2net}} & 2.6M & 1617.41  & 389.88  & 7.01 & 1.79 & & - & - & - & - \\
        \text{Ours} & 2.6M & 1150.27 & 281.94 & 3.84 & 1.20 & & 1095.26 & 280.42 & 3.53 & 1.19 \\
        \text{Ours w/S2D arch.} & 16.0M & 1211.97  &  296.19  & 4.24 & 1.33 & & - & - & - & -  \\
    
    \textbf{+ Self-Sup.}  &  &  & &\\
        \text{Ours+SS-S2D~\cite{ma2019self}} & 2.6M & \textbf{1112.83} & \textbf{268.79} & \textbf{3.27} & \textbf{1.12} & & \textbf{1062.48} & \textbf{268.37} & \textbf{3.12} & \textbf{1.13}\\
    \midrule \midrule
    \textbf{Supervised} & & & &\\
    
    \text{S-S2D~\cite{ma2019self}} & 27.8M & 878.56 & 260.90 & 3.25 & 1.34 & & 814.73 & 249.95 & 2.80 & 1.21 \\
    % eldesokey2018propagating
    \text{FusionNet~\cite{van2019sparse}} & 2.6M & 802.49 & 214.04 & 2.24 & 0.91 & & 772.87 & 215.02 & 2.19 & 0.93\\
    \text{DDP~\cite{yang2019dense}} & 18.8M & - & - & - & - & & 836.00 & 205.40 & 2.12 & 0.86\\
    % \text{DeepLiDAR~\cite{qiu2019deeplidar}} & - & 687.00 & 215.38 & 2.51 & 1.10 & & 758.38 & 226.50 & 2.56 & 1.15\\
    \bottomrule
\end{tabular}
\end{center}

\end{table}
\noindent\textbf{Impact of RGB Modality.} Contrary to self-supervised methods, which use RGB information to compute a photometric loss, we do not require the RGB image for good performance as shown in Table~\ref{tab:input_type}. Including RGB information reduces the error by 0.7\% in RMSE, and by using the CycleGAN RGB images the RMSE is reduced by 2.1\%. In a fully supervised manner the difference is 16.3\% for FusionNet \cite{van2019sparse}, showing that methods aiming to further reduce the RGB domain gap may increase the overall performance. Due to computational constraints, we train the CycleGAN model in a separate step. To test an end-to-end approach, we use the method in \cite{zheng2018t2net}, which does not use cycle-consistency, however we obtained lower-quality translated images and reduced accuracy as shown in Table~\ref{tab:input_type}.\par

\subsection{Method Evaluation}
\noindent\textbf{Comparison to State-of-the-Art.} In Table~\ref{tab:validation_comp} we compare our method, \textit{Ours}, with the real domain GT-free state-of-the-art. In the test set\footnote{Our entries are in the online leaderboard under the name \textit{SynthProj} for \textit{Ours} and \textit{SynthProjV} for \textit{Ours+SS-S2D}} our method decreases the RMSE by 6.4\%, the MAE by 6.3\%  and obtains better results for iRMSE and iMAE compared to VOICED~\cite{wong2020unsupervised}. Note that these improvements upon previous methods are obtained by using an architecture with fewer parameters. Table~\ref{tab:ablation_study} and Table \ref{tab:validation_comp} show that we achieve similar results to~\cite{wong2020unsupervised} by training only with synthetic data, i.e., in the first training step, which validates the observation that the main source of error to simulate are the see-through points. DDP~\cite{yang2018deep} uses synthetic ground truth from Virtual KITTI \cite{Gaidon:Virtual:CVPR2016} for training, however no adaptation is performed on the synthetic data, resulting in worse results compared to our method even when using stereo pairs (\textit{DDP+Stereo}). Both VOICED~\cite{wong2020unsupervised} and SS-S2D~\cite{ma2019self} use, besides video self-supervision, the noisy sparse input as supervision with no filtering, reducing the achievable performance as shown in Table \ref{tab:ablation_study} in \textit{No~Filter}. Non-learning methods in Table~\ref{tab:validation_comp} perform worse than~\cite{ma2019self,yang2019dense,wong2020unsupervised}. \par
% In \cite{yang2019dense}, Virtual KITTI~\cite{Gaidon:Virtual:CVPR2016} was used to learn the
% conditional prior. Virtual KITTI contains the same scenes as KITTI but they are computer-generated, hence the domain gap in scene information and RGB statistics is lower than for CARLA. We achieve better performance than~\cite{yang2019dense}, showing that adapting the sparse depth modality is key for good adaptation and more important than a lower domain gap in the RGB modality or the scenes presented.\par

\noindent\textbf{Domain Adaptation Baselines.} Following synthetic-to-real depth estimation methods \cite{atapour2018real,zheng2018t2net}, we use as a domain adaptation baseline a CycleGAN \cite{zhu2017unpaired} to adapt the images. To sparsify the synthetic depth, we use the real LiDAR masks \cite{atapour2019complete}, shown in Table~\ref{tab:ablation_study} to perform better than Bernoulli sampling. The performance of this domain adaptation baseline is presented in Table~\ref{tab:validation_comp} in \textit{DA Base}. We explore the use of adversarial approaches to match synthetic and real distributions on top of the \textit{DA Base}. \textit{DA Base + Disc. Out.} in Table~\ref{tab:validation_comp} uses an output discriminator using the architecture in \cite{pilzer2018unsupervised},
%which is widely used in semantic segmentation tasks \cite{tsai2018learning, hoffman2016fcns, chen2019learningsem}, 
with an adversarial loss weight of $0.001$ similarly to \cite{tsai2018learning}. Following \cite{zheng2018t2net}, we also tested a feature discriminator in the model bottleneck in \textit{DA Base + Disc. Feat.} with weight $0.01$. Table~\ref{tab:validation_comp} shows that the use of discriminators has a small performance impact and that standard domain adaptation pipelines are not capable of bridging the domain~gap.\par
\begin{table}[t]
\vspace{-2mm}
\caption{Semi-supervised results in the selected validation set for different pretraining strategies before finetuning on available annotations. \textit{S} and \textit{I} are the number of annotated sequences and images respectively. For \textit{Only supervised}, the weights are randomly initialized.}
\label{tab:semi_sup}
\begin{center}
\scriptsize
\begin{tabular}{lcccccccc} \toprule
& \multicolumn{2}{c}{\textbf{S:1 / I:196}} & & \multicolumn{2}{c}{\textbf{S:3 / I:1508}} & & \multicolumn{2}{c}{\textbf{S:5 / I:2690}}\\
 \cmidrule(lr){2-3}\cmidrule(lr){5-6}\cmidrule(lr){8-9}
        {Pretraining Strategy} & {RMSE} & {MAE} &\  \ \ & {RMSE} & {MAE} &\  \ \ & {RMSE} & {MAE} \\
    \midrule
        
        \text{Only Supervised} & 2578.72 & 1175.78 & & 1177.90 & 302.30 & & 1042.75 & 295.73 \\
        \text{DA Baseline}  & 1130.79 & 310.68 & & 1042.70 & 255.56 & & 986.09 & 244.94  \\
        \text{Ours} & \textbf{1106.30} & \textbf{262.29} & & \textbf{996.28} & \textbf{247.00} & & \textbf{949.63}  & \textbf{242.61}   \\
    \bottomrule
\end{tabular}
\end{center}

\end{table}

\noindent\textbf{Semi-Supervised Learning.} In some settings, a subset of the real data may be annotated. Our full pipeline mimics the noise in the real sparse depth and takes advantage of the unannotated data by using the filtered sparse depth $\mathcal{S}_p$ for supervision. This provides a good initialization for further finetuning with any available annotations as Table~\ref{tab:semi_sup} shows. Compared to pretraining using the \textit{DA Baseline}, our method achieves in all cases a better performance after finetuning.

\begin{figure}[t]
    \begin{center}
    \includegraphics[trim={0.22cm 0 0.8cm 0},width=0.32\columnwidth,clip]{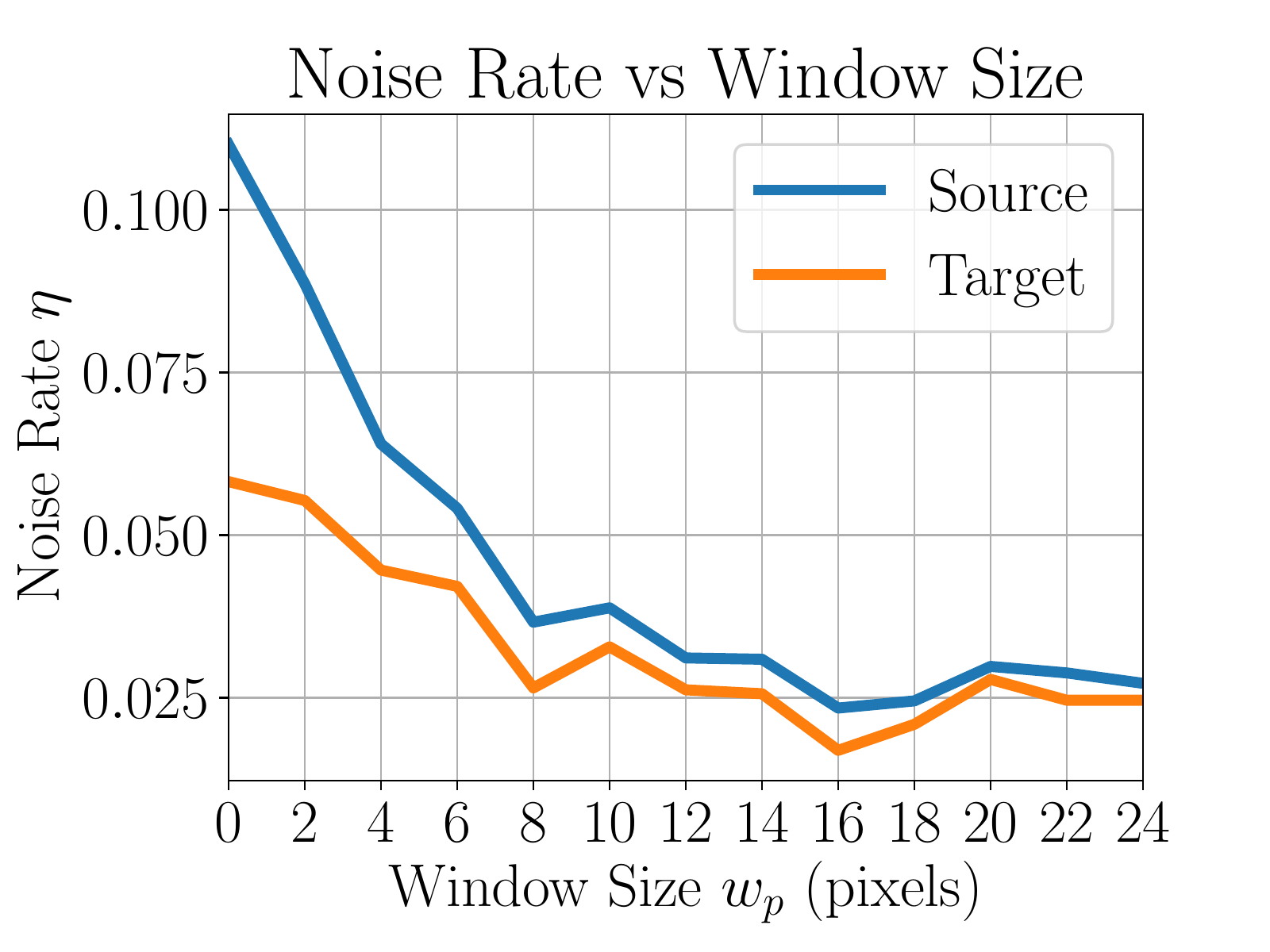}    \includegraphics[trim={0.22cm 0 0.8cm 0},width=0.32\columnwidth,clip]{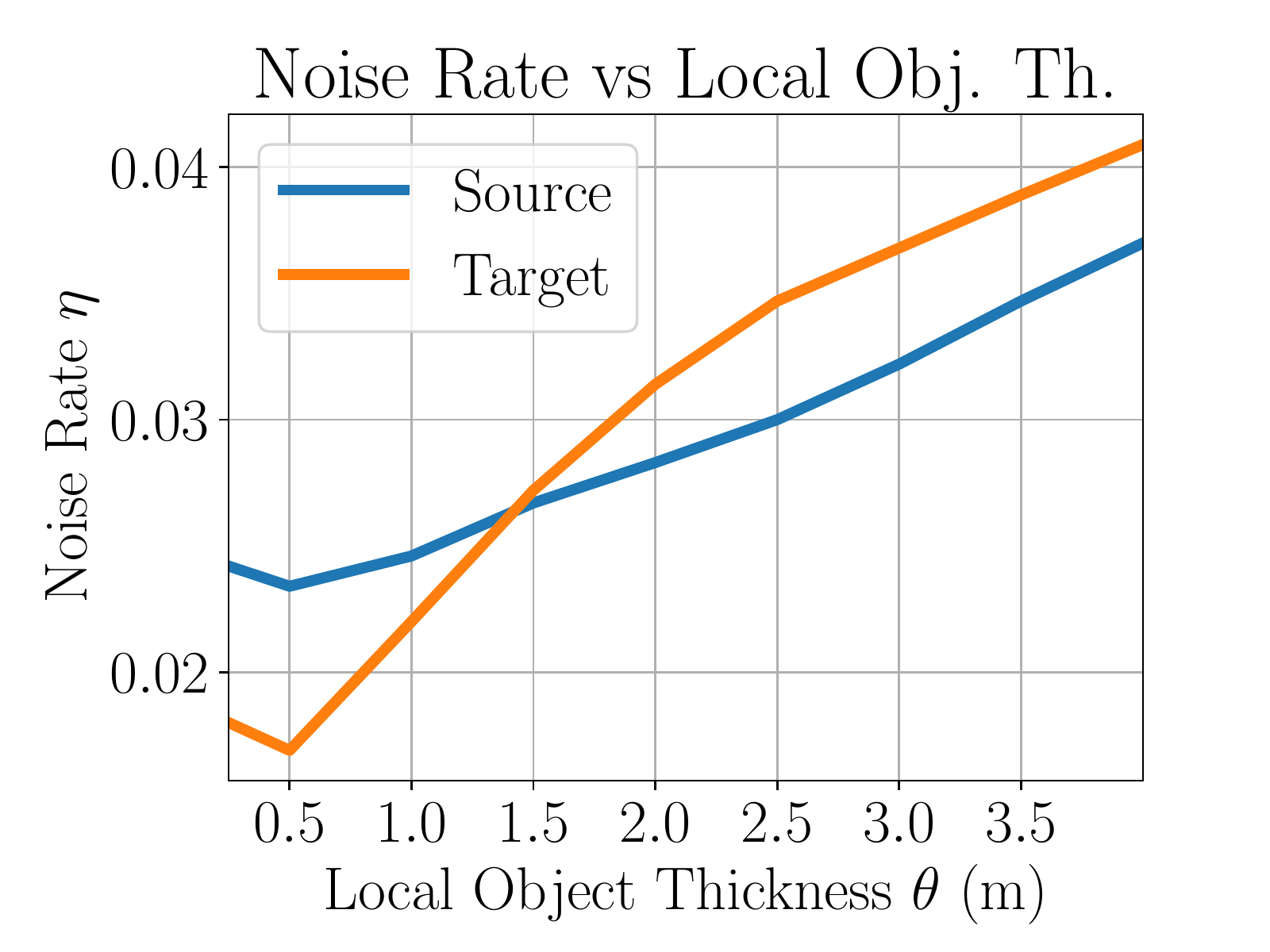}
        \includegraphics[trim={0.22cm 0 0.8cm 0},width=0.32\columnwidth,clip]{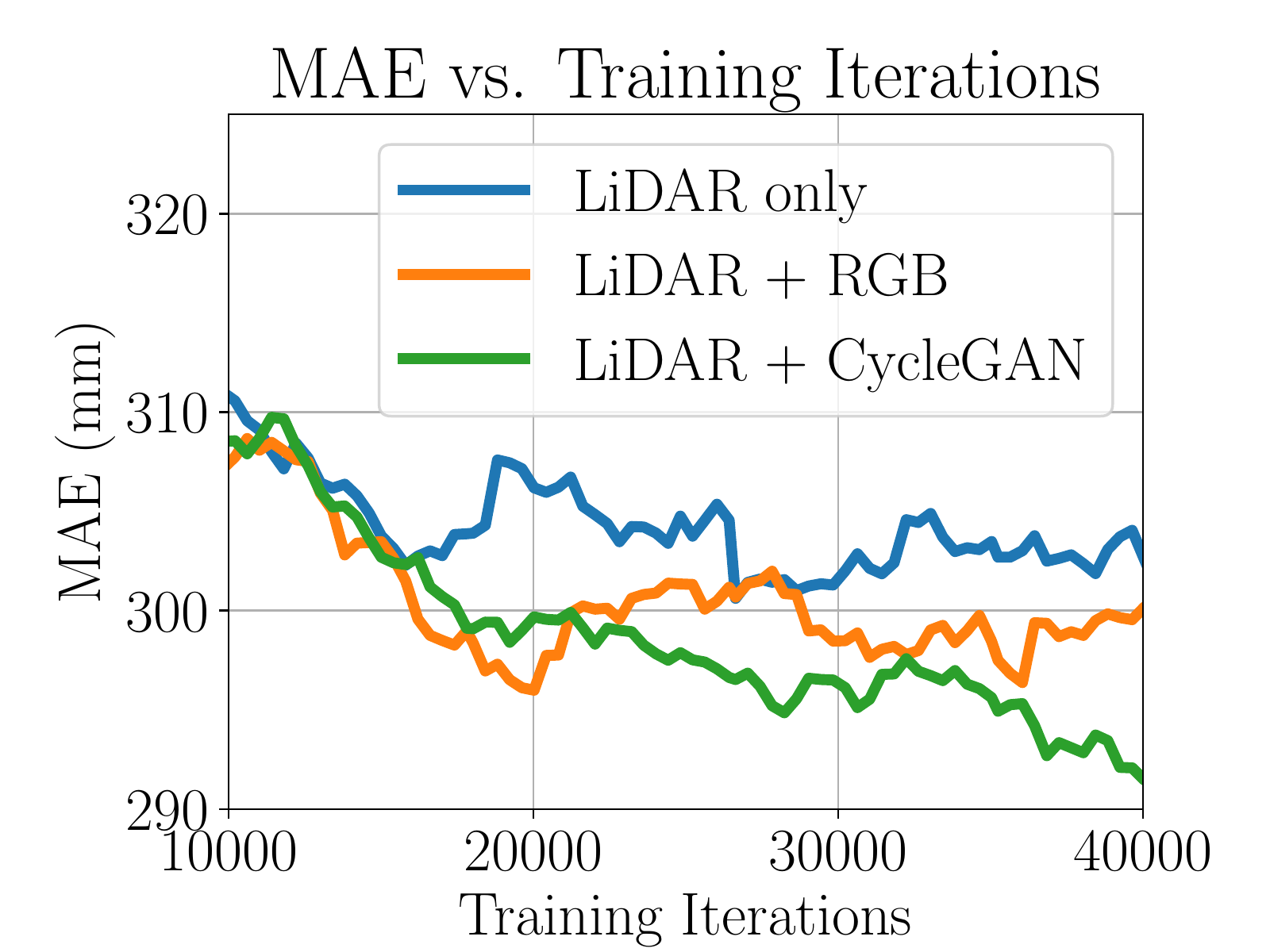}    
 \end{center}
 \vspace{-1.3em}
    \caption{Hyperparameter analysis. The two left images show the noise rate  $\eta$ vs. $w_p$ ($\theta=0.5$~m) and $\theta$ ($w_p=16$~pixels). The right plot shows MAE vs. number of training iterations in the second step, where we evaluate every 400 iterations, use a moving average with window size 25 and average 3 runs to reduce the variance}
    \label{fig:hyperparameter}
\end{figure}

\noindent\textbf{Hyper-Parameter Selection.} We do not tune $\lambda_S$ and $\lambda_R$ as we assume we do not have real-domain annotated data. The projected points $x^{s_P}_D$ in the synthetic validation set are used to choose $w_p$ and $\theta$ by employing $\eta$ in $\mathcal{S}_p$ as the indicator for the filtering process performance. We define a point to be noisy if the distance to the GT is bigger than 0.3~m, but other values tested yield similar results for the optimal $w_p$ and $\theta$. Figure~\ref{fig:hyperparameter} shows the noise percentage depending on $w_p$ and $\theta$, where we see that curves for $\eta$ follow a similar pattern in both the synthetic and real domain. We first select $w_p$ and then $\theta$ as the gain in performance is lower for $\theta$. The optimal values found are $w_p=16$~pixels and $\theta=0.5$~m. Figure~\ref{fig:hyperparameter} also shows the MAE depending on the number of iterations in the second step, where we empirically found that any chosen value yields better performance compared to using only synthetic data. After 40,000 training iterations, we did not see a noticeable improvement.\par

\begin{figure}[t]
    \begin{center}
    \includegraphics[trim={0cm 0cm 0 0},width=\linewidth,clip]{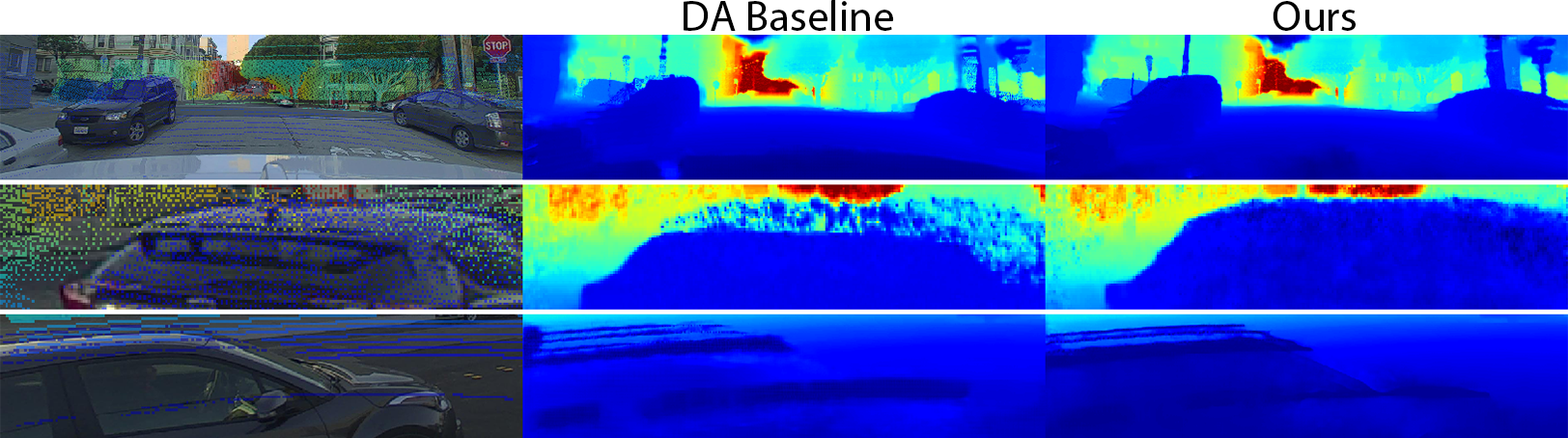}
    \end{center}
    \vspace{-1em}
    \caption{Qualitative results in PandaSet \cite{PandaSet} for both our \textit{DA Baseline} and full pipeline (\textit{Ours}) trained in CARLA and KITTI. The RGB image shows overlaid the sparse depth input. Despite the different camera set-up compared to the set-up used during training, our method is capable of correcting projection artifacts}
    \label{fig:pandaset}
\end{figure}

\noindent\textbf{Adding Self-Supervision.} When real domain video data is available, our approach can be combined with self-supervised methods~\cite{ma2019self,wong2020unsupervised}. \textit{Ours+SS-S2D} in Table~\ref{tab:validation_comp} adds the photometric loss $\lambda_{ph}\mathcal{L}_{ph}$ from~\cite{ma2019self} to our pipeline during the second step of training for the real data, with $\lambda_{ph}=10$ to have similar loss values as $\mathcal{L}_{S}+\mathcal{L}_{R}$. \textit{Ours+SS-S2D} further reduces the error in the test set and achieves, compared to VOICED~\cite{wong2020unsupervised}, a lower RMSE by 9.2\% and a lower MAE by 10.4\%.\par
\noindent\textbf{Model Agnosticism.} %Our proposed method is agnostic to the model used and we chose the FusionNet model~\cite{van2019sparse} due to its good performance and computational cost trade-off. 
We chose FusionNet~\cite{van2019sparse} as our main architecture, but we test our approach with the 18-layers architecture from~\cite{ma2019self} to show our method is robust to changes of architecture. Due to memory constraints we use the 18-layers architecture instead of the 34-layers model from \cite{ma2019self}, which accounts for the different parameter count in Table \ref{tab:validation_comp} between \textit{Ours w/S2D arch} and \textit{SS-S2D}. We set the batch size to 2, increase the number of iterations in the second step to 90,000 (the last 20,000 iterations use a lower learning rate of $10^{-4}$), and freeze the batch normalization statistics in the second step. The result is given in Table~\ref{tab:validation_comp} in \textit{Ours w/S2D arch.}, which achieves state-of-the-art RMSE and MAE.\par

\begin{figure}[t]
    \begin{center}
    \includegraphics[trim={0cm 0cm 0 0},width=\linewidth,clip]{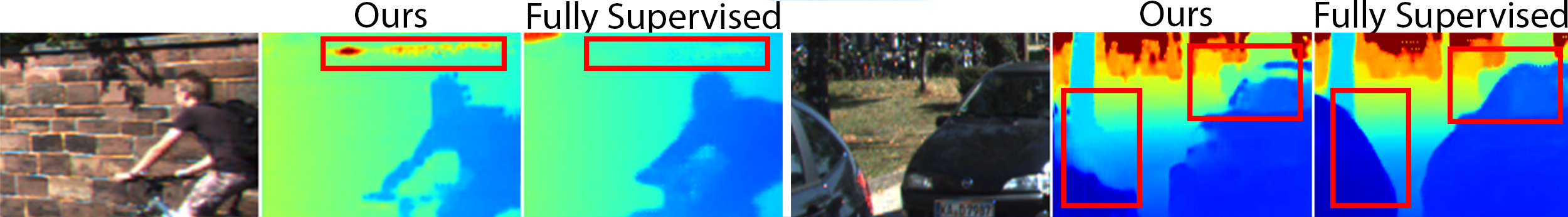}
    \end{center}
    \vspace{-1em}
    \caption{Failure cases of our method in KITTI, which cannot correct all types of noise. The left side example shows a set of noisy inputs on the wall. The right side example shows dropping of points due to low-reflectance black surfaces}
    \label{fig:limitations_examples}
\end{figure}
\noindent\textbf{Qualitative Results in PandaSet \cite{PandaSet}} are shown in
Figure \ref{fig:pandaset} for our full method compared to the DA Baseline (\textit{DA Base} in Table \ref{tab:validation_comp}) trained for CARLA and KITTI without further tuning. PandaSet contains a different camera set-up with different physical distances compared to the one used in training, e.g., top row in Figure \ref{fig:pandaset} corresponds to a back camera not present in KITTI. Our method is still capable of better correcting projection artifacts (top row and middle row) and completing the missing data (bottom row) compared to the DA Baseline.

\noindent\textbf{Limitations.} While we addressed see-through artifacts, other types of noise can be present in the real sparse depth as Figure~\ref{fig:limitations_examples} shows. The left side example shows a set of noisy inputs on the wall that is not corrected. The right side example shows missing points in the prediction due to the lack of data in the black hood surface. The fully supervised model deals properly with these cases, suggesting that approaches focused on other types of noise could further decrease the error.
%In the third row 
% \begin{figure}[h]
%     \centering
%     \includegraphics[width=\columnwidth,clip]{gfx/Errors_gt.jpg}
%     \caption{RGB image with overlayed semi-dense ground truth depth. Errors in the ground truth marked with a red rectangle. \adrian{Make points bigger}}
%     \label{fig:ground_truth_err}
% \end{figure}
% \vspace{-1.0em}
% \paragraph{Ground Truth Errors.} Annotating a real dataset with ground truth depth is challenging, especially in a depth completion setting where the performance is superior compared to RGB-only counterparts, hence the need for accurate ground truth is high.
% Due to this difficulty, the ground truth annotations are not completely dense, lacking information around the edges and also contain some annotation mistakes as shown in Figure~\ref{fig:ground_truth_err}.
% The inherent difficulty for reliably annotations make ground truth agnostic methods such as the one presented in this paper especially helpful.
% \noindent \textbf{Combining with self-supervision} We can use our method along with self-supervision approach used in past research. In this set-up, we assume we have both the stereo image pair and the calibration matrices, which are used jointly with the predicted depth to warp from one camera to the other. The warped image is compared to the original image to compute a photometric loss, which is the L1 loss in our case at multiple scales similarly to REF.

\section{Conclusions}
We proposed a domain adaptation method for sparse depth completion using data-driven masking and projections to imitate real noisy and sparse depth in synthetic data. The main source of noise in a joint RGB~+~LiDAR set-up was assumed to be the see-through artifacts due to projection from the LiDAR to the RGB reference frame. We also found a set of reliable points in the real data that are used for additional supervision, which helped to reduce the domain gap and to improve the performance of our model. A promising direction is to investigate the use of orthogonal domain adaptation techniques capable of leveraging the RGB inputs even more to correct also other types of error in the LiDAR co-modality.

\bibliographystyle{splncs}
\bibliography{project}

%this would normally be the end of your paper, but you may also have an appendix
%within the given limit of number of pages
\end{document}